\documentclass[10pt,twocolumn,letterpaper]{article}

\usepackage{iccv}
\usepackage{times}
\usepackage{epsfig}
\usepackage{graphicx}
\usepackage{amsmath}
\usepackage{amssymb}

\usepackage{bm}
\usepackage{url}
\usepackage{pifont}
\usepackage{bbding}
\usepackage{comment}
\usepackage{caption}
\usepackage{multirow}
\usepackage{makecell}
\usepackage{amsfonts}
\usepackage{booktabs}
\usepackage{subfigure}
\usepackage[misc]{ifsym}
\usepackage{pifont}
\usepackage[title]{appendix}
\usepackage{color}
\usepackage{xcolor,colortbl}
\newcommand{\cmark}{\ding{51}}%
\newcommand{\xmark}{\ding{55}}

\def\ie{{\em i.e. }}
\def\eg{{\em e.g. }}

\def\method{DeOP}
\def\TITLEMY{Open-Vocabulary}
\def\titlemy{open-vocabulary}

\usepackage[breaklinks=true,bookmarks=false]{hyperref}
\hypersetup{
    colorlinks=false,
    citecolor=black,
}

\makeatletter
\renewcommand{\paragraph}{%
  \@startsection{paragraph}{4}%
  {\z@}{0.5em}{-1em}%
  {\normalfont\normalsize\bfseries}%
}
\makeatother

\iccvfinalcopy 

\def\iccvPaperID{3421} 

\ificcvfinal\fi

\begin{document}

\title{{\TITLEMY} Semantic Segmentation with Decoupled One-Pass Network}

\author{Cong Han$^{1}\footnotemark[1]$
\quad Yujie Zhong$^{1}\footnotemark[1]$
\quad Dengjie Li$^1$
\quad Kai Han$^2\footnotemark[2]$
\quad Lin Ma$^1$\footnotemark[2] \\
$^1$Meituan Inc.
\quad $^2$The University of Hong Kong \\
{\tt\small hancong0911@163.com \qquad jaszhong@hotmail.com \qquad kaihanx@hku.hk}
}

\maketitle
\ificcvfinal\thispagestyle{empty}\fi

\renewcommand{\thefootnote}{\fnsymbol{footnote}}
\footnotetext[1]{Equal contribution.}
\footnotetext[2]{Corresponding authors.}

\vspace{-3mm}

\begin{abstract}
\vspace{-3mm}
Recently, the {\titlemy} semantic segmentation problem has attracted increasing attention and the best performing methods are based on two-stream networks: one stream for proposal mask generation and the other for segment classification using a pre-trained visual-language model. However, existing two-stream methods require passing a great number of (up to a hundred) image crops into the visual-language model, which is highly inefficient. To address the problem, we propose a network that only needs a single pass through the visual-language model for each input image. Specifically, we first propose a novel network adaptation approach, termed patch severance, to restrict the harmful interference between the patch embeddings in the pre-trained visual encoder. We then propose classification anchor learning to encourage the network to spatially focus on more discriminative features for classification. Extensive experiments demonstrate that the proposed method achieves outstanding performance, surpassing state-of-the-art methods while being 4 to 7 times faster at inference. Code: \href{https://github.com/CongHan0808/DeOP.git}{\textcolor{magenta}{https://github.com/CongHan0808/DeOP.git}}
\end{abstract}

\vspace{-5mm}
\section{Introduction}

Semantic segmentation is a critical computer vision task that entails grouping image pixels into semantically significant regions and predicting their class labels. Previously, semantic segmentation networks have concentrated on a pre-defined set of semantic classes based on the dataset. Recently, more attention has been devoted to {\titlemy} (also known as zero-shot) semantic segmentation, thanks to the rise of large-scale pre-trained visual-language models (VLMs) like CLIP~\cite{Radford2021CLIP} and ALIGN~\cite{Jia2021ALIGN}.

OpenSeg~\cite{Ghiasi2021OpenSeg} was among the first methods for the {\titlemy} semantic segmentation task, which proposes to adopt class-agnostic masks for possible semantic regions and then classify them using text embeddings extracted from a pre-trained VLM (ALIGN~\cite{Jia2021ALIGN} in its case).
It is efficient since it only requires passing the image through the visual encoder once for segmentation. \textbf{However, the single shared visual encoder that is updated during training inevitably breaks the original visual-language alignment of the pre-trained VLM.} Therefore, it requires training on numerous semantic concepts to ensure its zero-shot capability. For instance, training on Localized Narratives~\cite{PontTuset_eccv2020} is necessary. We refer to OpenSeg as the \emph{coupled one-pass} method. Another line of work, including ZegFormer~\cite{Ding2021ZegFormer}, SimBaseline~\cite{Xu2021SimpleBaseline} and OVSeg~\cite{Liang2022OpenVocabularySSOVSeg}, decouples the two sub-tasks and designs a two-stream architecture. On this note, the class-agnostic mask proposal stream and the mask classification stream use two separate backbones, preserving the generality of the VLM's visual encoder. \textbf{However, the downside of such approaches is their high computational overhead since image crops obtained by proposal masks are fed to the visual encoder individually.} We refer to methods like these as \emph{decoupled multi-pass} methods.

\begin{figure}[!t]
 \centering
 \subfigure[]{
  \begin{minipage}[b]{0.27\linewidth}   
   \includegraphics[width=\columnwidth]{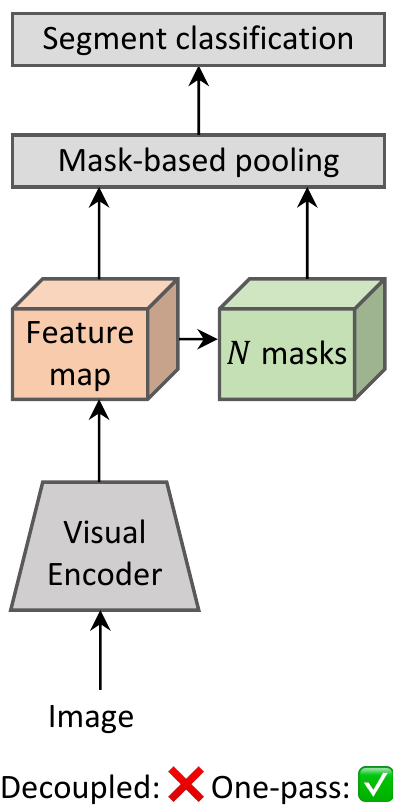}   
   \vspace{-5mm}
   \label{openseg}
  \end{minipage}
  }
  \subfigure[]{
  \begin{minipage}[b]{0.365\linewidth}  
   \includegraphics[width=\columnwidth]{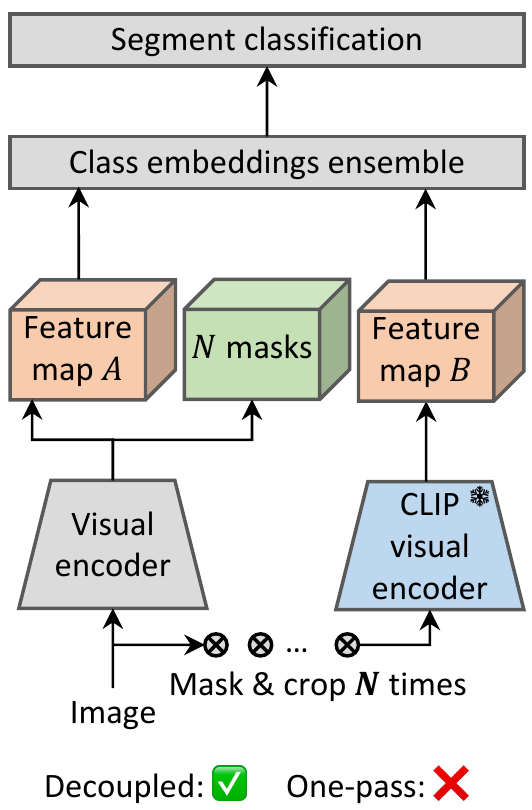}  
   \vspace{-5mm}
   \label{zegformer}
  \end{minipage}
  }
  \subfigure[]{
   \begin{minipage}[b]{0.27\linewidth}  
   \includegraphics[width=\columnwidth]{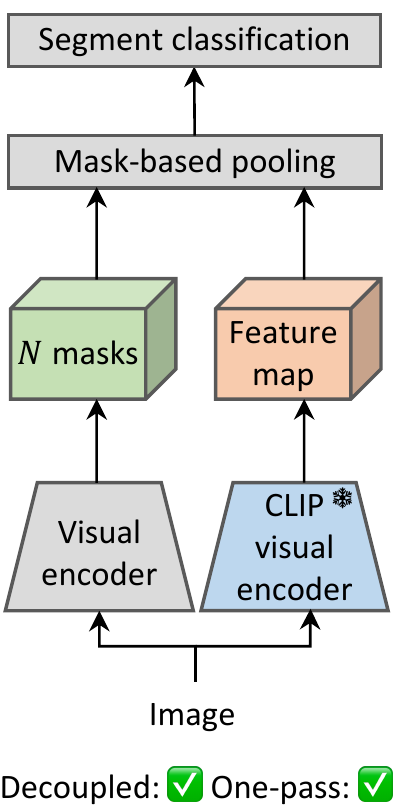}
   \vspace{-5mm}
   \label{baseline}
  \end{minipage}
 }
 \vspace{-2mm} 
 \caption{Comparisons between three macro-architectures for open-vocabulary semantic segmentation: (a) \textbf{coupled} network, \eg OpenSeg~\cite{Ghiasi2021OpenSeg}; (b) \textbf{decoupled multi-pass} network, \eg SimBaseline~\cite{Xu2021SimpleBaseline}; (c) \textbf{decoupled one-pass} network (our baseline).
 The decoupled structure can maintain the generality of the pre-trained VLM, while the one-pass mechanism brings high computational efficiency.} 
 \label{Overview} 
 \vspace{-2mm}
\end{figure}

In this work, we attempt to answer a question: can we develop a framework that can maintain the zero-shot ability of the VLM while being computationally efficient? To explore this possibility, we first design a baseline architecture based on two key design principles - \emph{decoupled} and \emph{one-pass}. 
As depicted in Figure~\ref{baseline}, our baseline approach features two decoupled visual backbones, while the visual encoder of the pre-trained VLM (\ie CLIP in this work) remains fixed to retain its generalization ability. Additionally, to achieve high efficiency, only one pass is required for the image in the classification stream.
However, we empirically find that the performance of this baseline architecture is far from ideal. We further apply prompt learning to adapt both the CLIP visual encoder and text encoder to the segmentation task at hand, but the performance gap between the baseline model and previous decoupled methods is still significant.

After conducting a thorough analysis, we discover that the main reason for poor classification performance lies in the segment classification rather than the quality/recall of the masks. Specifically, we have identified two main problems that contribute to this issue:
\textbf{(1) The patch embeddings that belong to different segments interact too much with each other in the original CLIP visual backbone.} In contrast, in multi-pass decoupled methods, patch embeddings in different masks have no interaction at all when passing through the VLM. Although patches belonging to different masks can provide some context information for classification, excessive interaction between different segment embeddings can actually harm the classification performance.
\textbf{(2) The final segment embedding, which is obtained by mask-based pooling of patch embeddings, is not optimal for classification.} This is because the mask proposal network is trained in a class-agnostic manner, and the weightings in the mask only indicate how likely each patch embedding belongs to a particular segment, regardless of the category. Therefore, pooling the patch embeddings based on these masks is sub-optimal for classification.
Therefore, by addressing these issues, we believe that it is possible to improve the classification performance of one-pass segmentation models while maintaining their computational efficiency.

To this end, we propose a two-stream framework, called Decoupled One-Pass network (\method), with the Generalized Patch Severance (GPS) and Classification Anchor Learning (CAL) to alleviate the above two problems, respectively.
\textbf{Generalized Patch Severance} can be seen as a `severance' operation on the patch tokens/embeddings in the CLIP visual encoder (\eg ViT-Base). It aims to reduce the harmful interference between patch tokens in the encoder, while maintaining the embedding space of CLIP.
\textbf{Classification Anchor Learning} is designed to find patches that are more suitable for segment classification, which we term classification anchors. It is achieved by appending a module at the end of the CLIP visual encoder, and the module learns to generate a heatmap for each mask proposal, indicating which patch embeddings should be focused (\ie the anchors) in the following  spatial pooling for classification. 

We extensively experiment on public benchmarks and show that \method~consistently outperforms previous methods in both intra- and cross-dataset evaluation, while being significantly more efficient than other multi-pass methods (e.g., SimBaseline), validating the effectiveness of our proposed GPS and CAL.

\vspace{-1mm}
\section{Related Work}
\vspace{-1mm}
\paragraph{Vision-language pre-training.} 
Vision-language pre-training aims to connect image-text representations for tasks such as image-text retrieval, visual question answering, text-to-image generation, and dense prediction. One prominent pre-trained model, CLIP~\cite{Radford2021CLIP}, learns image representations from a vast internet dataset of 400 million (image, text) pairs, delivering superior zero-shot image classification performance. To improve performance of pre-trained VLM for downstream tasks, text prompt learning has been proposed as a means to adapt models to new tasks by adding extra learnable tokens to the text prompt. Text prompt learning achieves strong generalization across various tasks in both few-shot and zero-shot settings, as demonstrated by studies such as \cite{Zhou2021PromptLearningTP,Ju2021PromptingVM,Feng2022PromptDetTO}.

\paragraph{Semantic segmentation.} Semantic segmentation is a fundamental task in computer vision used to cluster parts of an image that belong to the same category. Fully Convolutional Networks (FCNs)~\cite{Shelhamer2014FCN} represent a seminal work in deep-net-based semantic segmentation, formulating it as a per-pixel classification task. Various variants, such as ASPP~\cite{Chen2016DeepLabSI,Chen2017DeepLabV3}, PPM~\cite{Zhao2016PyramidSP}, and OCNet~\cite{Yuan2021OCNetOC}, have followed FCNs. In recent years, MaskFormer~\cite{Cheng2021MaskFormer,Cheng2021MaskedattentionMTMask2former} has proposed predicting a set of binary masks, each associated with a single global class label prediction.

\paragraph{Open-vocabulary semantic segmentation.} Open-vocabulary semantic segmentation learns pixel-wise classifiers for unseen object categories using either per-pixel-based (e.g., ZS3Net~\cite{Bucher2019ZS3Net}, CaGNet~\cite{Gu2020CaGNet}, SPNet~\cite{Xian2019SPNet}, Fusioner~\cite{Ma2022OpenvocabularySSXie} ) or region-based semantic segmentation (e.g. OpenSeg~\cite{Ghiasi2021OpenSeg}, ZegFormer~\cite{Ding2021ZegFormer}, SimBaseline~\cite{Xu2021SimpleBaseline}, OVSeg~\cite{Liang2022OpenVocabularySSOVSeg} and ODISE~\cite{xu2023odise}). ZS3Net uses a generative model while CaGNet incorporates context-aware feature generation. SPNet utilizes a fixed word embedding projection matrix. ZegFormer, SimBaseline as well as two concurrent works, OVSeg and ODISE, generate per-segment embeddings and then perform class-agnostic grouping and per-segment {\titlemy} classification. This work adopts a decoupled architecture similar to them. Notably, the decoupled architecture can be easily combined with more advanced networks (such as SAM-series~\cite{kirillov2023segany, zhao2023fastsam, zhang2023fastersam} or Mask DINO~\cite{li2023maskdino}) for better mask proposal generation.

\vspace{-2mm}
\section{Baseline Method}
\vspace{-1mm}
\begin{figure*}
\begin{center}
   \includegraphics[width=0.9\linewidth]{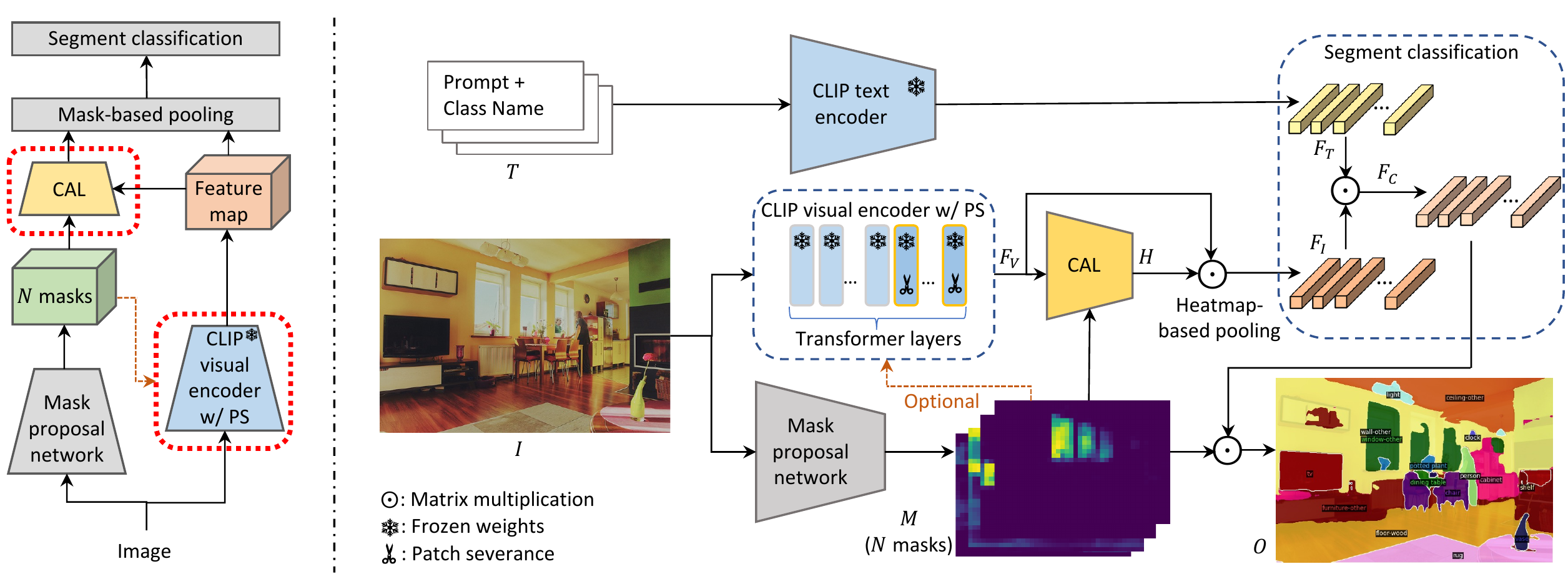}
\vspace{-3mm}
\end{center}
   \caption{\textbf{Left}: the overview of {\method}. Comparing to our baseline network, the additional patch severance (PS) and classification learning (CAL) module are indicated by red dotted line. 
   \textbf{Right}: the detailed pipeline. We feed an image to mask proposal network to generate $N$ proposal masks $M$, and the image also goes into the CLIP visual encoder (with PS) to obtain visual features $F_V$. $M$ and $F_V$ are passed to the CAL module to produce classification anchor heatmaps $H$ for $M$. $H$ is used in the heatmap-based pooling to obtain $F_I$. The text embeddings $F_T$ are generated by CLIP text encoder and used as classifiers for  to get the classification scores for each mask. Finally, we get the segment prediction by combining $F_C$ and $M$. 
   }
\label{pipline}
\vspace{-2mm}
\end{figure*}

\paragraph{Problem definition.}
We follow the settings of zero-shot segmentation in ZS3Net~\cite{Bucher2019ZS3Net}, which defines all classes as $C=C_{s}\bigcup C_{u}$ and $C_{s}\bigcap C_{u} = \varnothing$, where $C_s$ is the category set of seen classes and $C_u$ is the set of unseen classes. In training stage, the model needs to project the images into different regions with seen classes. During inference, the model predicts the results for all seen and unseen classes.

As discussed, we aim to tackle the {\titlemy} semantic segmentation task based on two key principles: one is to keep the generalization ability of the pre-trained VLM for unseen classes, and the other one is to achieve high computational efficiency. 
In the following, we first propose a baseline network which possesses the two desired abilities (Section~\ref{subsec:arch}), and then present an improved baseline method (\ie baseline$+$) which is obtained by applying prompt learning on the baseline (Section~\ref{subsec:prompt}). Lastly, in Section~\ref{subsec:limitation}, we discuss the limitations of the proposed baseline method.

\subsection{Baseline Network Architecture} 
\label{subsec:arch}
Figure~\ref{baseline} shows the decoupled one-pass architecture of the baseline method. 
The network contains two streams: a class-agnostic mask proposal network and a mask classification network.
The mask proposal network groups pixels to a series of class-agnostic masks. With this proposal network, we can convert the pixel-level classification to the region-level classification. 
The mask classification branch consists of a frozen CLIP visual encoder and classifiers obtained by the CLIP text encoder.

Specifically, we feed an image into the mask proposal network and CLIP visual encoder to generate $N$ proposal masks $M$ and the image feature map respectively. Then the $N$ proposal masks are applied to the CLIP feature map to obtain $N$ feature embeddings via mask-based pooling~\cite{Ghiasi2021OpenSeg}. 
On the other hand, the class embeddings $F_{C}$ (\ie the classifiers) are obtained by computing the dot product between the visual features $F_I$ with text the feature embedding $F_T$ generated by CLIP text encoder which encodes the text categories. Finally, we get the final prediction $O$ by combining the $N$ proposal masks with their class predictions using matrix multiplication. We can formulate these as $F_{C}=F_{T} * F_{I}$ and $O = F_{C} * M$.

\paragraph{Mask proposal network based on MaskFormer.} We build our mask proposal network by adapting MaskFormer~\cite{Cheng2021MaskFormer} into a class-agnostic variant. Unlike per-pixel classification segmentation models, MaskFormer is a mask classification model that predicts a set of masks, each associated with a single global class label.
To make MaskFormer class-agnostic, we simply remove the class embedding component while retaining the mask-related components. (Please refer to \ref{supp:maskformer} for a detailed description of the modified MaskFormer's structure).

\paragraph{Training of baseline.}
In our baseline method, only the mask proposal network needs training, whereas the classification branch requires no training. During training of the proposal network, only the mask loss is employed, and classification loss is removed. Additionally, we learn a text prompt in an offline manner, same as~\cite{Xu2021SimpleBaseline}.
\vspace{-1mm}
\subsection{Improving Baseline with Prompt Learning}
\vspace{-1mm}
\label{subsec:prompt}

Notably, since the CLIP visual encoder is frozen in order to retain its zero-shot ability, a straightforward approach to enhance the classification performance of the baseline is prompt learning~\cite{Zhou2021PromptLearningTP}. 
Prompt learning can slightly modify the embedding space of CLIP such that it can better adapt to the task at hand without breaking its visual-language alignment. 
In this work, we adopt both visual prompt tuning and text prompt learning to enhance our baseline, and the resultant method is denoted by baseline$+$.

\paragraph{Visual prompt tuning.} Visual Prompt Tuning~\cite{Jia2022VPT} (VPT) is an efficient and effective alternative to full fine-tuning for large-scale Transformer models in vision. We follow VPT to finetune the pre-trained CLIP visual encoder model. 
In specific, we add a set of learnable prompts to the sequence of the image patches embeddings in an element-wise manner, which can slightly steer the CLIP visual embedding space after training, so that it is more suitable for the segmentation task.

\paragraph{Online text prompt learning.}
In SimBaseline, the text prompts are learned offline via a classification task. To further improve the segmentation performance, we propose to train them with the whole segmentation network in an end-to-end manner.
Specifically, the learnable text prompts are first initialized offline as~\cite{Xu2021SimpleBaseline}, and then trained together with the visual prompts for the task of semantic segmentation.

\paragraph{Training of baseline$+$.}
The baseline$+$ requires end-to-end training to learn the visual and text prompts for the CLIP encoders. As it is trained for segmentation, we adopt a combination of dice loss~\cite{Li2019DiceLoss} and focal loss~\cite{Lin2017FocalLF}, which is a  common practice for semantic segmentation.

\subsection{Limitations}
\label{subsec:limitation}

Experimentally, the performance gap between the baseline$+$ approach and previous decoupled approaches (\ie ZegFormer and SimBaseline) remains substantial, as shown in Table~\ref{ablation}. It can be deduced that the gap is mainly caused by the segment classification, since their mask proposal networks are similar. In this work, we consider two problems that lead to the inferior classification performance: (1) The original CLIP visual encoder may not be aware of the correspondence between patch embeddings and semantic segments, and the patch embeddings belonging to different segments may overly/aimlessly interact in the encoder. Notice that this problem does not exist in previous decoupled methods since the patch embeddings belonging to different segments are not interacted, owing to the separate pass through the visual encoder of masked images. (2) The final segment embeddings produced by mask-based pooling on the patch embeddings are not ideal for classification. The mask proposal network is trained in a class-agnostic manner; therefore, the output masks may not be suitable for mask-based pooling for segment classification.
\vspace{-2mm}
\section{Our Method -- \method}
To mitigate the two problems discussed in Section~\ref{subsec:limitation}, we propose a ViT-adapting approach coined patch severance to reduce the harmful interaction between patch embeddings in the CLIP visual encoder, and the Classification anchor learning (CAL) to produce better  representations in the mask-based pooling for classification.

\paragraph{Overall pipeline.}
An overview of our framework, \method, which follows the two-stream design of our baseline method, is shown in Figure~\ref{pipline} (left). The two additional components, patch severance and CAL, are highlighted with red dotted lines.
The proposed patch severance is applied to one or more layers in the CLIP visual encoder, while CAL is achieved by appending a query-based heatmap decoder after the CLIP visual encoder.
The rest of the pipeline (demonstrated in Figure~\ref{pipline} right), including the mask proposal network, classifiers generated from text and segment classification are the same as the baseline network.
In terms of the training of \method, the same loss as baseline$+$ is adopted.
Notably, at inference, \method~does not need a heuristic ensemble of the classification scores from the classification stream and the mask proposal stream as required in~\cite{Ding2021ZegFormer,Xu2021SimpleBaseline}.

In the following, we elaborate on the proposed patch severance in Section~\ref{subsec:sps} and classification anchor learning in Section~\ref{subsec:cal}.

\subsection{Patch Severance}
\label{subsec:sps}

Generally, the feature maps extracted by the ViT-based CLIP visual encoder contain a large amount of global information of the whole image, due to the self-attention layer in the transformer layers. However, such property may be overwhelming for the semantic segmentation task. 

The classification branch aims to classify each region. Although a certain extent of information from other regions may provide some context and benefit the classification, too much information may have negative effects and distract the classification. 
 
To this end, we propose a ViT adaptation approach, termed \textbf{patch severance} (PS), to restrict the interference between patch embeddings in the CLIP visual encoder. 
In the following, we compare two types of patch severance: mask-guided patch severance and generalized patch severance.
They can be applied flexibly in one or more layers in the ViT-based visual encoder.

\begin{figure}[!t] 
 \centering
 \includegraphics[width=0.9\linewidth]{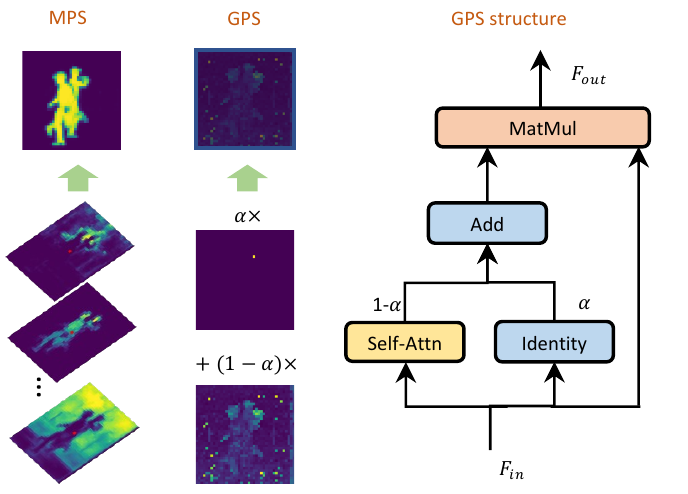}

\caption{
\textbf{Left}: the comparison between  MPS and GPS in terms of how to obtain the final attention heatmap (for a particular patch embedding) in the self-attention layer. 
\textbf{Right}: the structure of GPS, which is used to replace the original self-attention layer in the transformer layer in ViT.
}
\label{SPS}
\vspace{-2mm}
\end{figure}
 
\paragraph{Mask-guided patch severance.}
Mask-guided Patch Severance (MPS) is an operation to guide the interaction between patch embeddings in the transformer layer. 
As Figure~\ref{SPS} (left) illustrates, MPS replaces the multiplication between key and query in the self-attention operation by proposal masks (generated by the mask proposal network), such that each patch embedding mainly interacts with those within the same predicted segment, while having little interaction between those belonging to other predicted segments.
The mask-guided attention for each patch embedding is derived from the $N$ proposal masks.
Concretely, the mask-guided attention for the $i$th patch embedding is computed by Equation~\ref{mps}:

\vspace{-3mm}
\begin{equation}
    M_i = \sum_{j=1}^{N} M_{j,i} * M_j,
    \label{mps}
    \vspace{-2mm}
\end{equation}
where $M_i$ is the mask attention map for the $i$th patch embedding, $M_j$ is the $j$th mask and $M_{j,i}$ is the value of mask $M_j$ for $i$th patch embedding.
\paragraph{Generalized patch severance.}
Different from MPS which utilizes proposal masks as a strong prior to guide the patch interaction, a Generalized Patch Severance (GPS) is proposed and it requires no prior (Figure~\ref{SPS} middle).

GPS replaces the query-key multiplication in the self-attention with a weighted combination of the query-key multiplication and an identity matrix.
Mathematically, GPS can be expressed by Equation~\ref{gps}, where $E$ is an identity matrix, $Attn$ is the self-attention~\cite{Vaswani2017AttentionIA} matrix.
In one extreme, GPS degrades to the normal self-attention when $\alpha=0$. 
In the other extreme, when $\alpha=1$, GPS is equivalent to complete patch severance (\ie no interaction happens in the self-attention layer).
The hyper-parameter $\alpha$ enables a transition between the two extreme cases.
\vspace{-3mm}
\begin{equation}
    F_{out} = (\alpha * E + (1-\alpha)* Attn) \times F_{in}.
\vspace{-2mm}
\label{gps}
\end{equation}
As shown in Figure~\ref{pipline}, the patch severance can be applied multiple times to different transformer layers, while in practice we found that simply applying it to the last transformer layer is an effective choice. 
\vspace{-3mm}
\subsubsection{Comparison to Existing Adaptation Methods}

\paragraph{The embedding space is preserved.}
Previous adaptation methods like ViT-adapter~\cite{Chen2022ViT-Adapter} usually require an updating of the backbone network, which could easily break the original embedding space.
However, similar to VPT~\cite{Jia2022VPT}, the proposed patch severance can preserve the original visual embedding space of the pre-trained VLM (\eg CLIP), such that the visual-language alignment and the zero-shot generality can be maintained. This property is in line with our first design principle.
\paragraph{Training-free.}
Unlike existing methods (such as ViT-adapter~\cite{Chen2022ViT-Adapter} and VPT~\cite{Jia2022VPT}) which introduce additional parameters to the backbone network and thus requires further finetuning, patch severance brings no additional parameters and is training-free. 
Moreover, patch severance can work collaboratively with previous adapting methods (\eg VPT) to achieve better adaptation for the task at hand.
\subsection{Query-Based Classification Anchor Learning}
\label{subsec:cal}
Classification Anchor Learning (CAL) is proposed to overcome the second limitation, namely, focusing on the more discriminative patches for classification (\ie classification anchors) .
For example, to classify a human, the network may focus more on the head compared to the rest of the body.
Essentially, CAL shares a similar spirit with the anchor selecting/learning for training object detectors~\cite{zhang2019freeanchor,zhang2020bridging,feng2021tood}, while it aims to output an anchor heatmap for each proposal mask.
To achieve this, we propose a query-based heatmap decoder module. 
As Figure~\ref{pipline} shows, the heatmap decoder (indicated by CAL) is appended to the CLIP visual encoder.
It can generate attention heatmaps indicating which patches are more appropriate to represent the segments in terms of classification. The module is trained end-to-end with the whole segmentation network. We elaborate on it in the following.

\begin{table*}[t!]
\begin{center}
\resizebox{1\linewidth}{!}{ 
\begin{tabular}{lcccccccccccc}
\toprule
Method&Decoupled&Passes &&\multicolumn{4}{c}{COCO-Stuff} && \multicolumn{4}{c}{VOC} \\
\cmidrule(rl){5-8}
\cmidrule(rl){10-13}
 & &&&hIoU & Seen & \textbf{Unseen} & FPS && hIoU & Seen & \textbf{Unseen} & FPS\\
\hline
ZS3Net~\cite{Bucher2019ZS3Net} & \xmark & 1 && 15.0 & 34.7 & 9.5 &- && 28.7 & 77.3 & 17.7 & -\\
CaGNet~\cite{Gu2020CaGNet} & \xmark & 1 && 18.2 & 35.5 & 12.2 &-&& 39.7 & 78.4 & 25.6 &-\\
ZegFormer~\cite{Ding2021ZegFormer} & \cmark & $N^{'}$ && 34.8 & 36.6 & 33.2 &-&& 73.3 & 86.4 & 63.6&- \\
SimBaseline~\cite{Xu2021SimpleBaseline} & \cmark & $N^{'}$ && 37.8 & \textbf{39.3} & 36.3&1.11 &&77.5 & 83.5 & 72.5& 2.66\\
\method~(ours) & \cmark & 1 && \textbf{38.2} & 38.0 & \textbf{38.4} & \textbf{4.37} &&  \textbf{80.8} & \textbf{88.2} & \textbf{74.6} & \textbf{6.55} \\
\bottomrule
\end{tabular}
}
\end{center}
\vspace{-2mm}
\caption{Comparison with other methods on COCO-Stuff and Pascal VOC in the {\titlemy} setting. \emph{passes} means the number of passing of the input image through the CLIP visual encoder. 
Note that $N^{'}$ is dependent on the input image since ZegFormer and SimBaseline only pass the image crops with scores above a threshold into CLIP, and $N^{'}\le{N}$.}
\vspace{-2mm}
\label{intradataset}
\end{table*}
\vspace{-3mm}

\subsubsection{Anchor Heatmap Decoder}
We first describe the proposed query-based heatmap decoder, and then introduce a simple counterpart based on convolution. Both of them can achieve CAL and they are compared experimentally in Section~\ref{sec:exp}.
\begin{figure}[!t] 
 \centering
\includegraphics[width=0.9\linewidth]{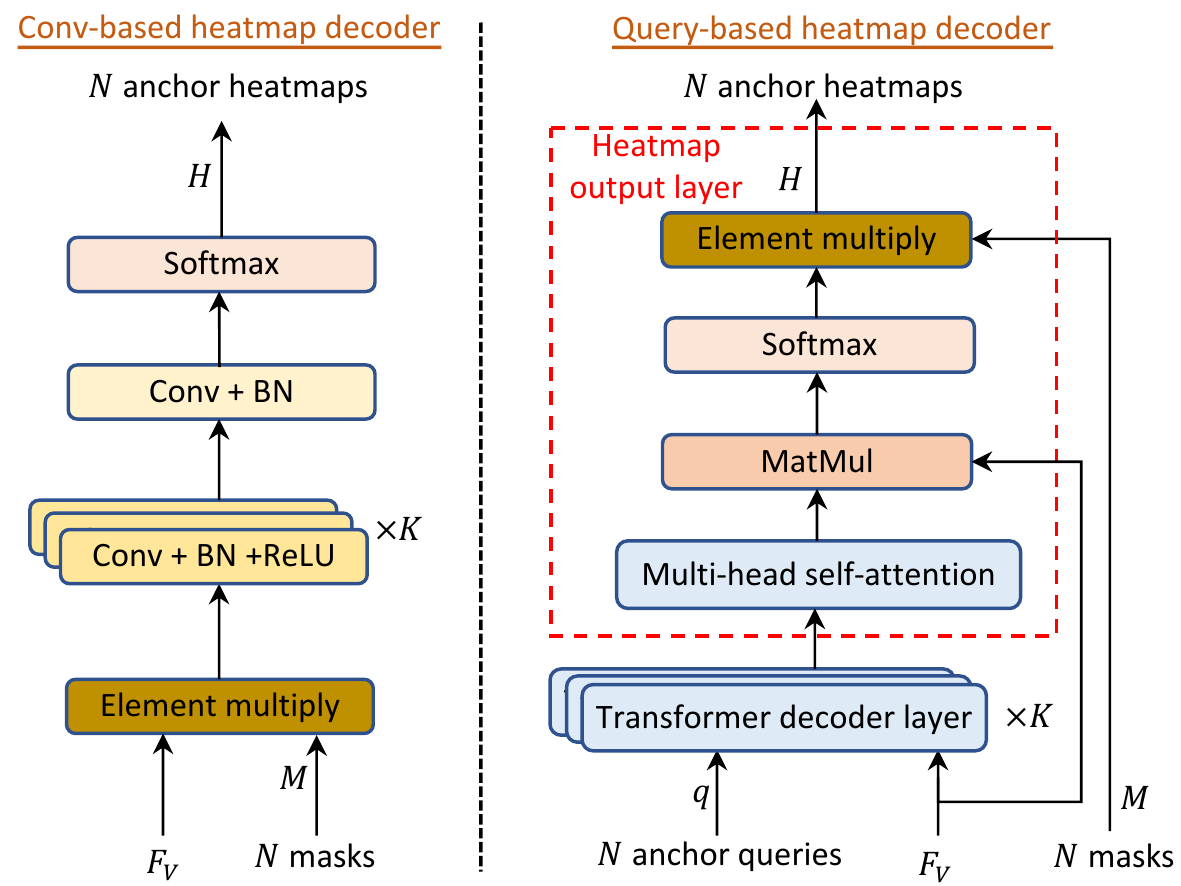}
\caption{
Comparison between the conv-based heatmap decoder and the novel query-based heatmap decoder for CAL.
}
\label{CAL}
\end{figure}
\paragraph{Query-based heatmap decoder.}

Overall, the decoder receives $N$ learnable anchor queries (denoted by $q$), feature maps output by CLIP visual encoder (denoted by $F_V$) and $N$ proposal masks (denoted by $M$). The output is $N$ classification anchor heatmaps $H$. 
The structure of the query-based heatmap decoder is displayed in Figure~\ref{CAL}(right).
The decoder is composed of $K$ standard transformer decoder layers and a novel heatmap output layer. 
The $K$ transformer layers are designed to update the anchor queries $q$ by attending to the feature map of the input image. 

The key component in the heatmap decoder is the \textbf{heatmap output layer}.
Unlike the standard transformer decoder layer in which the output is the updated queries, the proposed heatmap decoder layer is designed to produce anchor heatmaps $H$. 
In specific, the anchor queries first go through a multi-head self-attention layer, and then are multiplied with the feature map followed by Softmax to obtain an attention map for each anchor query. Lastly, the attention maps are multiplied with the $N$ proposal masks, enabling each anchor query to be aware of its corresponding mask/segment.
Concretely, the heatmap output layer can be formulated as Equation~\ref{attnquery}:
\vspace{-2mm}
\begin{equation}
    H =softmax(\frac{q\times F_V}{\sqrt{d}}) \ast M,
    \label{attnquery}
    \vspace{-1mm}
\end{equation}

\noindent where $q \in \mathbb{R}^{B\times N\times C}$ is the $N$ anchor queries, $F_V \in \mathbb{R}^{B\times C \times H \times W}$ is the feature map generated by CLIP visual encoder, $d$ is the dimension of $F_V$, $M \in \mathbb{R}^{B \times N \times H \times W}$ is $N$ masks and $\ast$ denotes the element-wise multiplication.

\paragraph{Heatmap decoder based on convolution.} 
The proposed query-based heatmap decoder is not the only way to achieve CAL. In this work, we also compare the query-based heatmap decoder with a convolutional counterpart (shown in Figure~\ref{CAL} left). 
This conv-based anchor heatmap decoder first generates $N$ weighted feature maps by applying each proposal mask to the feature map. Each weighted feature map goes into a stack of $K$ conv-BN~\cite{Ioffe2015BatchNABN}-ReLU~\cite{Agarap2018DeepLURELU} layers, where conv layers have kernel size of 3 and channel size 512). Lastly, an extra conv layer maps the channel dimension to 1, followed by a Softmax layer to generate the $N$ anchor heatmaps.
Mathematically, the conv-based anchor heatmap decoder can be expressed as follows:
\vspace{-1mm}
\begin{equation}
    H= \sigma(conv(CBR_{K} (\cdots  CBR_{1} ( F \ast M )))),
    \label{convdecoder}
    \vspace{-1mm}
\end{equation}
\noindent where $CBR$ refers to the conv-BN-ReLU layers, and $\sigma$ denotes Softmax operation.

\vspace{-3mm}
\subsubsection{Classification Based on Anchor Heatmap}
\vspace{-2mm}
Here we describe how to make use of the classification anchor heatmap for the  segment classification.
We simply follow the mask-based pooling described in~\cite{Ghiasi2021OpenSeg}, while the only difference is that the learned classification anchor heatmap is used for the pooling, instead of the mask proposal. We term this process heatmap-based pooling.
It can be formulated as $F_{I}=H \odot F_{V}$,
where $F_I$ is the classification embeddings for the proposal masks/segments.

\vspace{-3mm}
\subsubsection{Discussion on CAL}
\vspace{-2mm}
The proposed CAL can keep the original visual-alignment of CLIP because it only learns a spatial heatmap which down-weights or up-weights certain patch embeddings output by CLIP. 
Furthermore, although CAL is trained in seen classes, it can be well-transferred to unseen classes. This impressive property is verified quantitatively in Section~\ref{sec:exp}.

\vspace{-2mm}
\section{Experiments}
\vspace{-2mm}
\begin{table*}[t]
\begin{center}
\begin{tabular}{lcccccccccc} 
\toprule
Method& Backbone & Training dataset & De. & Passes & VOC-20 & PC-59 & A-150 & PC-459 & A-847 \\
\hline
ZS3Net~\cite{Bucher2019ZS3Net} & R101 & Pascal VOC 2012& \xmark & 1 & 38.3 & 19.4 & - & - & -\\ 
LSeg~\cite{Li2022LanguagedrivenSSLSeg} & R101& Pascal VOC 2012 & \xmark &1& 47.4& - & - & - & - \\
OpenSeg~\cite{Ghiasi2021OpenSeg} & R101 & COCO & \xmark & 1 &60.0 & 36.9 & 15.3 & 6.5 & 4.0 \\
OpenSeg~\cite{Ghiasi2021OpenSeg} & R101 & COCO + Loc. Narr. & \xmark & 1 &63.8 & 40.1 & 17.5 & 7.9 & 4.4  \\ 

SimBase~\cite{Xu2021SimpleBaseline} &R101c&COCO-Stuff-156 & \cmark & $N^{'}$ &88.4 & 47.7 & 20.5 & - & 7.0 \\
\method~(ours) & R101c & COCO-Stuff-156& \cmark & 1 & \textbf{91.7} & \textbf{48.8} & \textbf{22.9} & \textbf{9.4} & \textbf{7.1}\\
\bottomrule
\end{tabular}
\end{center}
\vspace{-2mm}
\caption{Comparison with other methods on Pascal VOC, Pascal Context and ADE20K in the cross-dataset setting. The number after each dataset indicates the number of categories. $De.$ means decoupled network. $Loc.\ Narr.$ means Localized Narratives~\cite{PontTuset_eccv2020}. 
}
\vspace{-1mm}
\label{cross-dataset}
\end{table*}

\begin{table*}[t!]
\begin{center}
\resizebox{0.8\linewidth}{!}{ 
\begin{tabular}{lccccccccc}
\toprule
Method &\multicolumn{4}{c}{COCO-Stuff} && \multicolumn{4}{c}{VOC} \\
\cmidrule(rl){2-5}
\cmidrule(rl){7-10}
 & pAcc &hIoU & Seen & \textbf{Unseen}& & pAcc & hIoU & Seen & \textbf{Unseen} \\
\hline
Baseline & 15.6 & 7.0 & 6.2 & 8.0  & &46.6 & 30.7 & 37.5 & 26.0\\
Baseline$+$ & 27.6 & 8.3 & 8.0 & 8.7 && 64.3 & 33.4 & 53.2 & 24.4\\
Baseline$+$ w/ PS & 53.9 & 25.8 & 25.9 & 25.4 && 72.7& 47.6& 64.0 & 37.9 \\
Baseline$+$ w/ CAL & 54.2 &23.2 & 28.8 & 21.0 &&88.1& 70.5& 83.3 &61.2\\
\method~(ours) & \textbf{62.2} & \textbf{38.2} & \textbf{38.0} & \textbf{38.4} &&\textbf{92.5}&\textbf{80.8} & \textbf{88.2} & \textbf{74.6} \\
\bottomrule
\end{tabular}
}
\end{center}
\vspace{-2mm}
\caption{Ablation on the effectiveness of PS (\ie patch severance) and CAL (\ie classification anchor learning).}
\vspace{-2mm}
\label{ablation}
\end{table*}

\paragraph{Datasets.} We study {\method} using four widely used image segmentation datasets: COCO-Stuff~\cite{Caesar2016COCOStuffTA,Lin2014MicrosoftCCCOC}, PASCAL VOC 2012~\cite{Everingham10VOC}, PASCAL Conext~\cite{mottaghi_cvpr14Context} and ADE20K~\cite{zhou2017sceneADE20k,zhou2019semanticADE20K}.
In terms of data splitting and evaluation metrics, we simply follow~\cite{Xu2021SimpleBaseline}. The details regarding dataset statistics, data splitting and evaluation metrics are described in the supplementary material.

\paragraph{Implementation details.} 

We conduct experiments based on Detectron2~\cite{wu2019detectron2}. We select ResNet-101c~\cite{chen2017deeplab} which replaces the first $7 \times 7$ convolution layer of ResNet-101~\cite{He2015DeepRLResNet} with $3$ consecutive $3 \times 3$
convolutions as the backbone for MaskFormer, and set the number of queries $N$ (\ie proposal masks) in the transformer decoder to 100 by default. For CLIP, we use the text encoder and image encoder of the ViT-B/16~\cite{Dosovitskiy2020AnIIViT} model.
The experiments are conducted on 8 Tesla V100 32G GPUs, and the batch size is set to 32. The training iterations are 20k for PASCAL VOC 2012 and 60k for COCO-Stuff. For the speed test, we test on COCO-Stuff dataset with batch size 1, using 1 Tesla V100 32G GPU.

\vspace{-2mm}

\subsection{Results in Open-Vocabulary Setting}
\vspace{-2mm}
We evaluate our method in a {\titlemy} segment setting by training the model on the COCO-Stuff dataset and the PASCAL VOC 2012 dataset. During training, we only use images belonging to seen classes in the training dataset, while including all images in the validation dataset.
Table~\ref{intradataset} shows the comparison with other methods.

As shown in Table~\ref{intradataset}, {\method} significantly outperforms previous works. Impressively, we achieve 38.2 hIoU for the COCO-Stuff dataset and 80.8 hIoU for the PASCAL VOC 2012 dataset.
The performance boost over ZegFormer and SimBaseline is mainly due to the proposed patch severance and learned anchor heatmaps for classification. Apart from that, we posit that those multi-pass methods get rid of the context (\ie information around the segment) completely for each proposal segment (due to the mask operation in the pre-processing) which may hinder the performance; whereas {\method} utilizes context information naturally in the one-pass architecture.

\paragraph{Inference speed.}
To evaluate the efficiency of our method, we compare the inference speed between our method and SimBaseline in Table~\ref{intradataset}. 
Impressively, our method attains around 3 to 4$\times$ increase in the inference speed (FPS). This demonstrates the benefits and efficiency of adopting the one-pass mechanism, which is our second design principle.

\vspace{-1mm}
\subsection{Results in Cross-Dataset Setting}
\vspace{-1mm}
We evaluate our method in a cross-dataset setting by training the model on the COCO-Stuff dataset and evaluating it on other datasets without fine-tuning. Table \ref{cross-dataset} shows that our approach generalizes well on other datasets due to the decoupled design (\ie the first design principle) and frozen visual CLIP encoder, and outperforms all previous methods. Specifically, our method achieves 91.7 mIoU on the Pascal VOC dataset, 48.8 mIoU on the Pascal Context dataset, and 22.9 mIoU on the ADE20K dataset. It is worth noting that as the number of categories increases from left to right, the corresponding performance metrics decrease, demonstrating the importance of improving classification.

\paragraph{Effectiveness of prompt learning, PS and CAL.}
\label{section:eachpart}
In Table~\ref{ablation}, we validate the effectiveness of each of our proposed components over the plain \emph{baseline}.
Prompt learning is benefiting the pAcc metric, which increases from 15.6 to 27.6 on COCO-Stuff dataset and from 46.6 to 64.3 on VOC dataset, as shown on \emph{baseline$+$}. Both PS and CAL  bring  improvements in all metrics. Not surprisingly,  \method (containing the prompt learning, PS and CAL) achieves a significant improvement over all others, demonstrating that PS and CAL can work collaboratively and generalize well to unseen categories.  In the supplementary material, we provide a detailed ablation on different combinations of text and visual prompt learning.

\begin{table}
\begin{center}
\begin{tabular}{lcccccc}
\toprule
Method & $\alpha_{-2}$ &$\alpha_{-1}$ & hIoU & Seen & \textbf{Unseen} \\
\hline
Baseline & 0 & 0  & 7.0 & 6.2  & 8.0 \\
MPS & - & - & 12.5 & 11.0 & 14.4 \\
GPS & 0 & 0.5& 20.0 & 17.1 & 24.0 \\
GPS   & 0& 1.0 & \textbf{23.8} & \textbf{21.5} & \textbf{26.6} \\
GPS  & 0.5& 1.0  & 16.2& 15.2 & 17.4 \\
GPS & 0.5 & 0.5 & 13.3 & 12.1 & 14.8 \\
\bottomrule
\end{tabular}
\end{center}
\vspace{-2mm}
\caption{Ablation study on patch severance. $\alpha_{-1}$ and $\alpha_{-2}$ denote the hyper-parameter $\alpha$ for the last layer and second last layer in ViT, respectively. Experiments are based on our baseline method (without prompt learning).}
\vspace{-2mm}
\label{AbaltionSPS}
\end{table}

\begin{figure*}[!t] 
 \centering
   \includegraphics[width=0.95\linewidth]{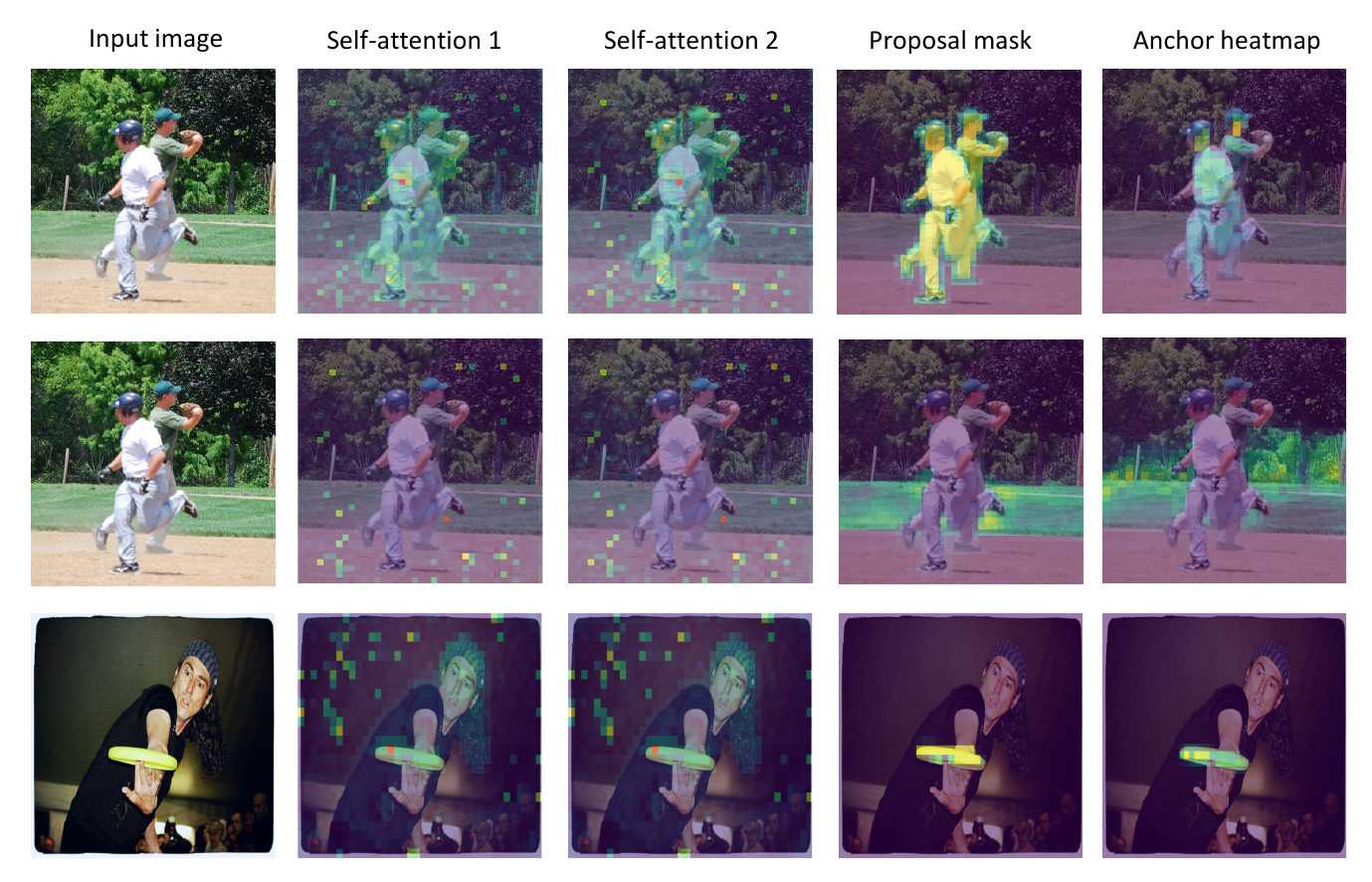}   
  \caption{The visualization of the heatmaps. From left to right are: the original image, the self-attention heatmap for a particular patch $i$ in the last transformer layer in the CLIP visual encoder, the self-attention heatmap for another patch $j$, the proposal mask for the segment containing the patch $i$ and the learned classification anchor heatmap for the patch $i$. 
  The categories in the three rows are: human (seen), grass (\textbf{unseen}) and frisbee (\textbf{unseen}).
  } 
 \label{VisualAttn} 
\end{figure*}

\vspace{-2mm}
\paragraph{Analysis on PS.} 
As discussed in Section~\ref{subsec:sps}, we compare two types of patch severance: MPS and GPS. The results are shown in  Table~\ref{AbaltionSPS}.
The first conclusion to draw is that both types of PS perform better than the original self-attention.
In general, GPS achieves better performance than MPS. We think that it is because the prior (\ie weighted proposal masks) is too strong and may not be optimal.
Interestingly, the best combination is that the last layer has a complete patch severance.
Notably, we also attempt to apply PS to more transformer layers in ViT. 
However, we find that PS may break the embedding space if it is applied in shallower layers (\ie the first layer) and results in poor performance which is shown in Table \ref{AbalationGPS}.

\begin{table}
\begin{center}
\begin{tabular}{lccccc}
\toprule
Type& $K$ & pAcc & hIoU & Seen & \textbf{Unseen} \\
\hline
 {\method} w/o CAL& 0 & 53.9 & 25.7 & 25.9 & 25.4 \\
Conv-based &3 &58.8 & 31.4 & 31.4 & 31.4 \\
Query-based & 1 & \textbf{62.2} & \textbf{38.2} & \textbf{38.0}  & \textbf{38.4} \\
Query-based & 3 &62.2& 37.7 &37.9& 37.6 \\
Query-based & 5 & 62.2 & 38.0 & 38.0 &  38.0\\
\bottomrule
\end{tabular}
\end{center}
\vspace{-2mm}
\caption{Ablation study on the classification anchor learning. 
$K$ is the number of layers in the heatmap decoder.
}
\vspace{-2mm}
\label{AblationCALDesign}
\end{table}
\vspace{-2mm}
\paragraph{Analysis on CAL.} 
We compare the query-based and conv-based heatmap decoder for CAL introduced in Section~\ref{subsec:cal}. 
For the query-based one, we also investigate the effect of the depth ($K$) of the decoder.
The results in Talbe~\ref{AblationCALDesign} illustrate that query-based heatmap decoder is better than the conv-based counterpart, and the performance is not sensitive to the depth of the heatmap decoder.

\paragraph{Results on larger CLIP visual encoder.}
\vspace{-1mm}
We also study the impact of the more powerful CLIP visual encoder on the outcomes. Specifically, we train SimBaseline and {\method} models with a CLIP visual encoder based on ViT-L/14-336 on the COCO-Stuff dataset. As displayed in Table ~\ref{tab:vitL}, our approach consistently outperforms SimBasline on unseen classes. Moreover, {\method} achieves a more impressive (almost 7$\times$) speed-up at inference.  
\begin{table}[!t]
\vspace{-1mm}
    \centering
    \begin{tabular}{lccccc}
    \toprule
    Method  & pAcc & hIoU & Seen & \textbf{Unseen} & FPS \\
    \hline
     SimBaseline  & 62.1 & 38.6 & \textbf{40.7} & 37.0 & 0.27 \\
     \method~(ours)  & \textbf{62.6} & \textbf{40.0} & 40.4& \textbf{39.6} & \textbf{1.85} \\
    \bottomrule
    \end{tabular}
    \caption{Results with CLIP visual encoder based on ViT-L.}
    \vspace{-2mm}
    \label{tab:vitL}
\end{table}
\vspace{-2mm}
\subsection{Visualization}
\label{sunsec:visualization}
\vspace{-3mm}
\paragraph{Comparison in attention masks.}
We visualize the attention maps, mask proposals, and anchor heatmaps in Figure~\ref{VisualAttn}. It is clear that the self-attention map for a patch is disorganized, and cannot provide useful information for the segmentation task. 
The attention map generated by CAL focuses on the local information surrounding the patch (\eg attending to the tree to help for classifying the grass), 
and CAL also allows the model to pay more attention to the locations that are more discriminative for classification, \eg the head being more important than the body of the human.
More heatmap visualizations and segmentation results are displayed in the supplementary material.

\label{sec:exp}

\vspace{-2mm}
\section{Conclusion}
\vspace{-1mm}
\vspace{-1mm}
In this work, we propose an efficient framework called {\method} for {\titlemy} semantic segmentation. Unlike previous decoupled methods, {\method} only requires a single pass for the input image in the visual-language model. Specifically, {\method} is equipped with the proposed patch severance, which restricts redundant interference between patch embeddings in the visual encoder of the visual-language model, and classification anchor learning to identify visually discriminative regions for better segment classification. We conducted extensive experiments to validate the method and found that {\method} outperforms previous methods in both intra- and cross-dataset evaluation while being significantly faster than other multi-pass methods during inference.
\paragraph{Acknowledgements}
This work is supported by National Key R\&D Program of China (No. 2022ZD0118700),
Hong Kong Research Grant Council - Early Career Scheme (Grant No. 27208022), and HKU Seed Fund for Basic Research. 
\vspace{-2mm}
{\small
\bibliographystyle{ieee_fullname}
\bibliography{egbib}
}

\clearpage
\appendix
\appendixpage 

\section{Class-agnostic Mask Proposal Network}
\label{supp:maskformer}
\begin{figure}[t]
\begin{center}
   \includegraphics[width=0.95\linewidth]{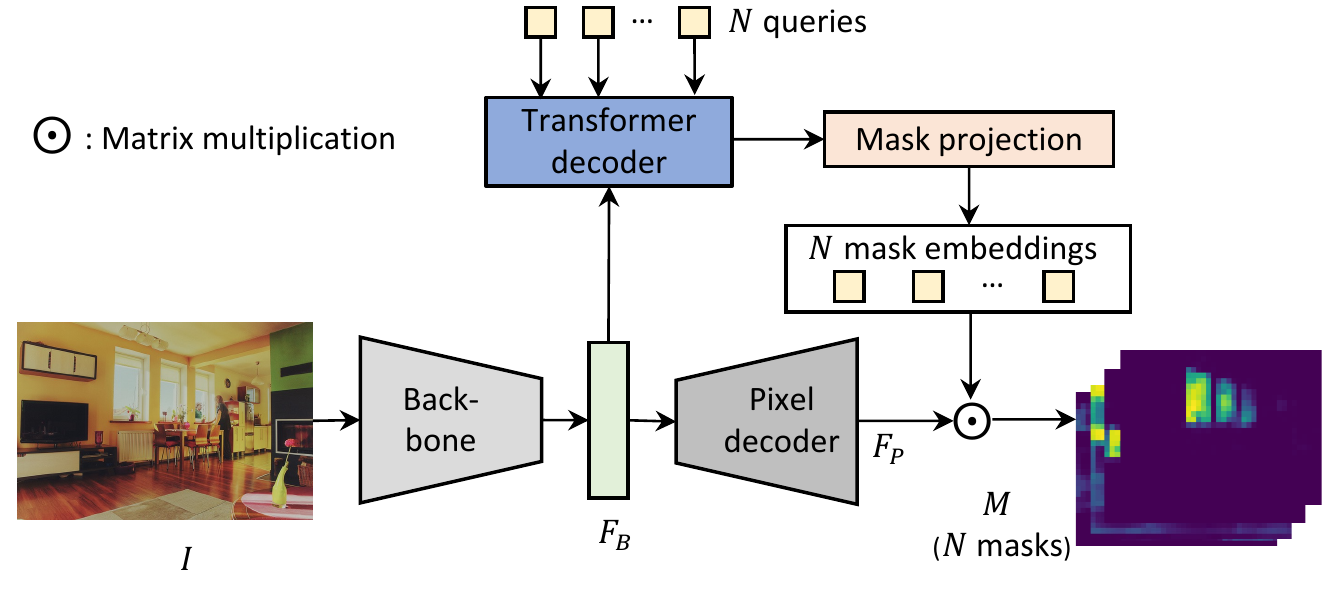}

\end{center} 
\caption{The architecture of class-agnostic mask proposal network.}
\label{figure:masknet}
\end{figure}

We provide a detailed illustration of the mask proposal network in this section.
We utilize a modified MaskFormer~\cite{Cheng2021MaskFormer} as our mask proposal network, which is mentioned in Section~\ref{subsec:arch} of the main paper. We present the overall framework of our class-agnostic mask proposal network in Figure~\ref{figure:masknet}. The model architecture is essentially the same as MaskFormer, while the classification branch is removed. 
We use a backbone to extract image features $F_B$. The image features are fed into a pixel decoder and a transformer decoder to generate $N$ mask embeddings and per-pixel embeddings $F_{P}$. Finally, we combine the $N$ mask embeddings and $F_{P}$ using matrix multiplication to get masks $M$. The quality of the proposal masks is assessed in Section~\ref{supp:proposal}.

\section{Visual Prompt Learning}
\begin{figure}
\begin{center}
   \includegraphics[width=0.9\linewidth]{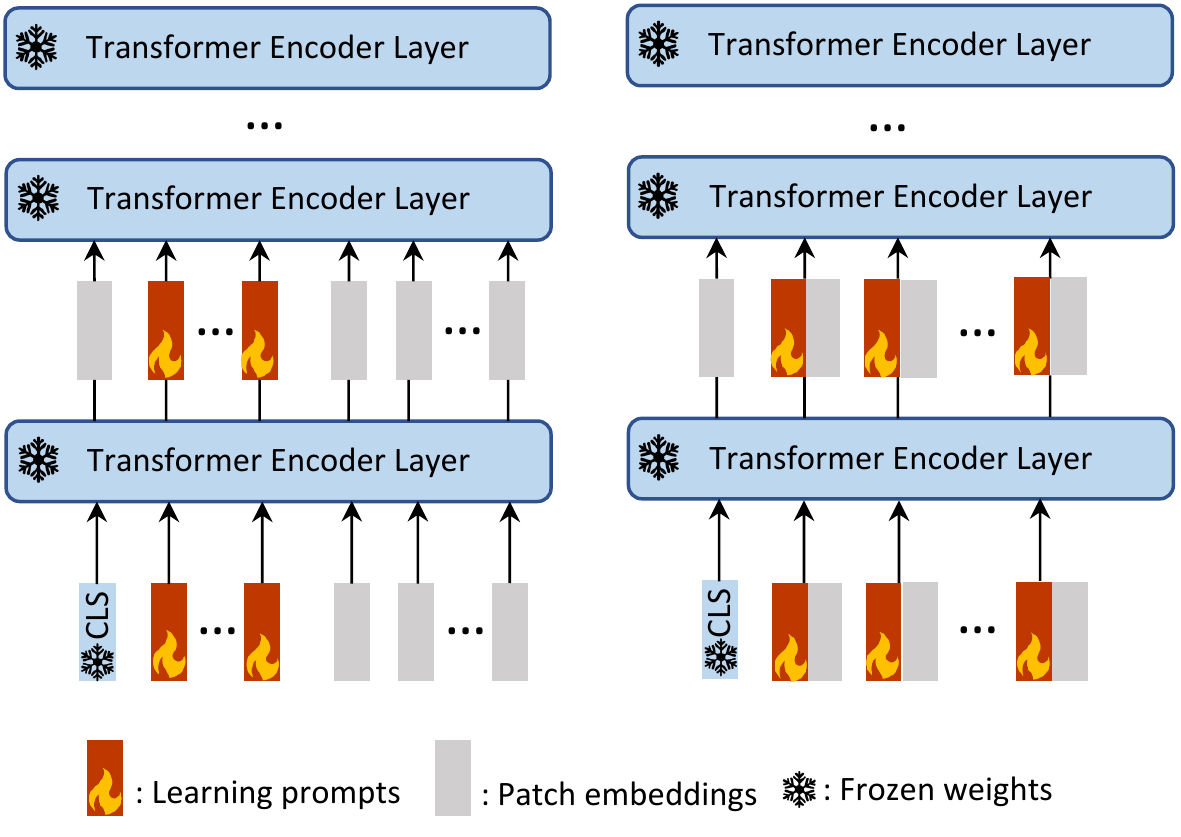}

\vspace{-3mm}
\end{center} 
\caption{Overview of visual prompt tuning. There are two implementations: prepending prompts (\textbf{Left}) and adding prompts (\textbf{Right}).}
\label{figure:vpt}
\vspace{-2mm}
\end{figure}

As discussed in Section~\ref{subsec:prompt} in the main paper, we improve the performance of the baseline method by fine-tuning the pre-trained model with prompt learning. The text prompt learning is explained in detail in the main paper, so here we elaborate on the visual prompt learning that we adopted. The architecture of visual prompt learning is demonstrated in Figure~\ref{figure:vpt}. We compare two forms of visual prompt learning: prepending prompts and adding prompts. The difference between these two forms lies in how the prompts are combined with image patch embeddings. $Prepending\ prompts$ refer to prepending $P$ prompt embeddings before $O$ image patch embeddings, resulting in $P+O$ embeddings, together with a class embedding. $Adding\ prompts$ involves adding a prompt embedding to each image patch embedding in an element-wise manner, and the length of prompt embeddings should be the same as the length of image patch embeddings.

In this section, we first provide the details of the datasets and evaluation metrics, and then provide further analysis of our methods, by including the ablation study on different prompt learning approaches, the effectiveness analysis (of the main components) on Pascal VOC dataset, and the performance of the mask proposal network.

\subsection{Datasets and Evaluation Metrics}
\paragraph{Datasets.}
\textbf{COCO-Stuff} is a large dataset for semantic segmentation that span over 171 categories including 80 things, 91 stuff. It contains 117k training images and 5k validation images. \textbf{PASCAL VOC} contains 11,185 training images and 1,449 validation images from 20 classes. \textbf{PASCAL Context} is a set of additional annotations for PASCAL VOC 2010. It contains 4,998 training images and 5,005 validation images. We select a subset of 59 frequent classes for use. \textbf{ADE20K} contains more than 20K scene-centric images for training and 2k images for validation. There are totally 150 semantic categories, which include stuff and discrete objects. 
\section{Experiment}

\paragraph{Data split.} We choose two types of data splits for validating our method on zero-shot semantic segmentation (ZS3) setting and cross-dataset setting respectively. For ZS3 setting, we follow the class split in \cite{Bucher2019ZS3Net}. 
In particular, on COCO-Stuff, we choose 156 classes as the seen classes and the rest 15 classes as the unseen testing classes. On Pascal VOC 2012, we choose 15 classes as the seen classes and the rest 5 classes as the  unseen testing classes. 
For cross-dataset setting, we train the model on COCO-Stuff seen classes dataset and validate on other datasets.

\begin{figure*}
\begin{center}
   \includegraphics[width=0.9\linewidth]{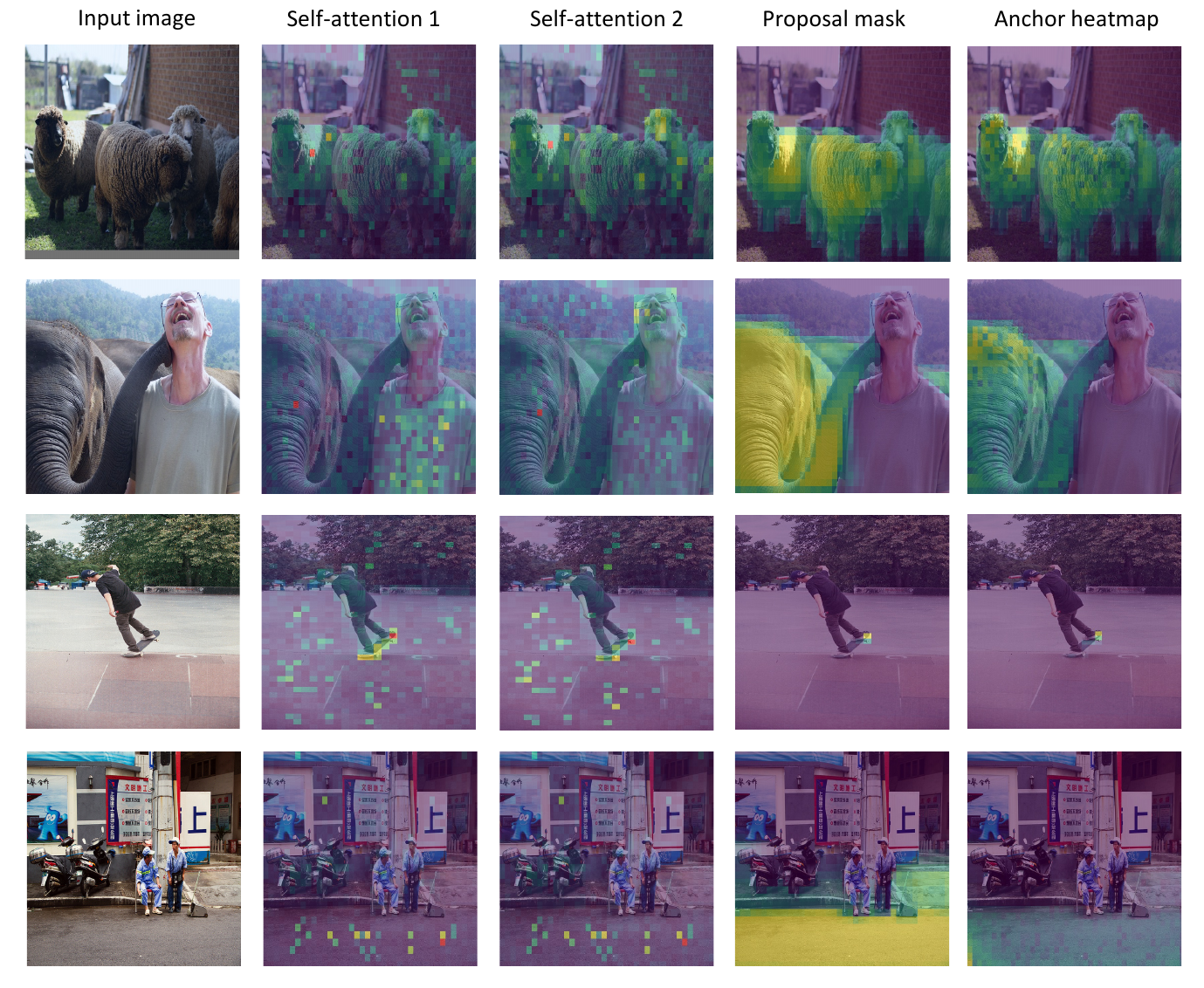}
\vspace{-3mm}
\end{center} 
\caption{The visualization of the heatmaps. The three categories in the four rows are: sheep (seen), elephant (seen), skateboard (\textbf{unseen}) and road (\textbf{unseen}).}
\label{figure:segment}
\vspace{-2mm}
\end{figure*}

\paragraph{Evaluation metrics.} Following the previous work, we measure pixel-wise classification accuracy (pAcc) and mean IoU (mIoU) for seen and unseen classes denoted as mIoU(S) and mIoU(U) respectively. Additionally, we compute the harmonic mean IoU (hIoU) among seen and unseen classes by the previous works \cite{Xian2019SPNet}, which is calculated as $hIoU = \frac{2*mIoU(S)*mIoU(U)}{mIoU(S) + mIoU(U}$.
For cross-dataset validation, we use mIoU as the evaluation metric.

\subsection{Prompt Learning}
\paragraph{Analysis on prompt learning.}

\begin{table}[!t]
\begin{center}
\resizebox{0.9\linewidth}{!}{ 
\begin{tabular}{lccccc}
\toprule
Type & End-to-end &pAcc & hIoU & Seen & \textbf{Unseen} \\
\hline
Text & \xmark & 15.6 & 7.0 & 6.2 & 8.0  \\
Text &\cmark & 25.7 & 7.2 & 7.6 & 6.9\\
Vision A. & \cmark& 22.6& 6.4 & 6.3 & 6.5 \\
Vision P. & \cmark & 16.7 &6.7 & 6.4 & 7.4 \\
Both & \cmark & \textbf{27.6} & \textbf{8.3} & \textbf{8.0} & \textbf{8.7} \\
\bottomrule
\end{tabular}
}
\end{center}
\caption{Ablation study on Prompt Learning. $Vision\ P.$ and $Vision\ A.$ mean prepending and adding prompts for visual prompt tuning. $Both$ refers to combining end-to-end text prompt learning and adding visual prompt learning.}
\label{AblationPrompt}
\end{table}
\begin{figure*}[!t] 
 \centering
   \includegraphics[width=0.9\linewidth]{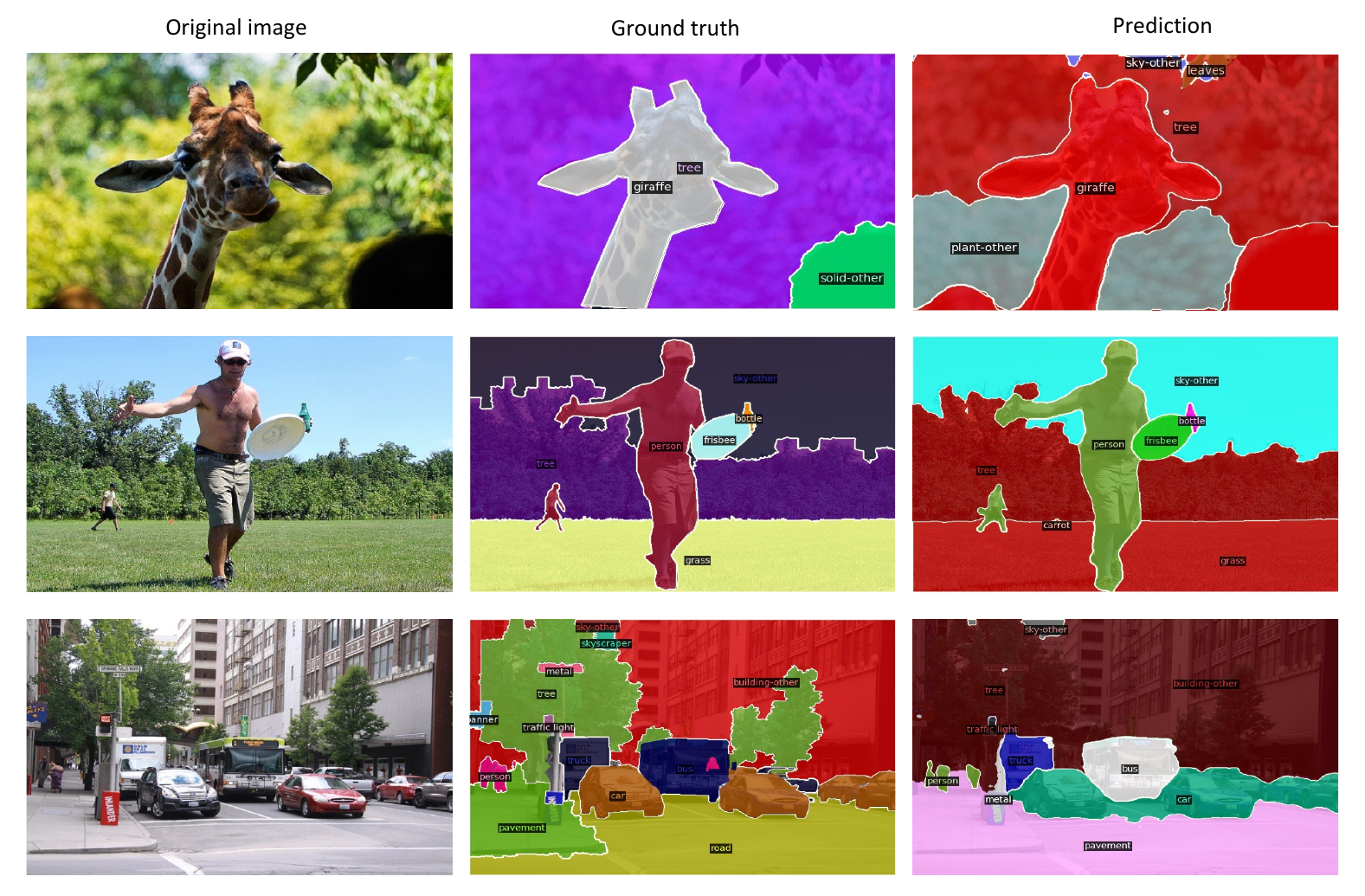}   
  \caption{The visualization of semantic segmentation. From left to right:  origin images,  ground truth semantic segmentation maps, and  predictions.} 
 \label{fig:VisualSegment} 
\end{figure*}
In Table~\ref{AblationPrompt}, we evaluate the effectiveness of different prompt learning approaches for segmentation task building upon the baseline method.
In Section~\ref{subsec:prompt} of the main paper, we propose improving the baseline method with prompt learning. To evaluate the impact of prompt learning, we conduct experiments with different types of prompts, based on the baseline method. We can observe that all prompt learning approaches can enhance model performance. The best results are achieved when combining end-to-end text prompt learning with adding visual prompt.

\begin{table}[t]
\begin{center}
\begin{tabular}{lcccc}
\toprule
Method & All & Seen & \textbf{Unseen}  \\
\hline
Recall{@}30 & 0.59 & 0.57 & 0.71  \\
Recall@50 & 0.45 & 0.44 & 0.56  \\
\bottomrule
\end{tabular}
\end{center}
\vspace{-2mm}
\caption{Recall of mask proposals on COCO-Stuff dataset. $Recall@30$ and $Recall@50$ means recall at IoU $ 30\%$ and $50\%$ respectively.}
\label{table:recallmask}
\end{table}

\begin{table}
\begin{center}
\begin{tabular}{lccccc}
\toprule
Method & layers  & hIoU & Seen & \textbf{Unseen} \\
\hline
Baseline & -  & 7.0 & 6.2  & 8.0 \\
GPS & [1] & 4.6 & 4.5 & 4.7 \\
GPS   & [1-6] & - & 0.0 & 0.1 \\
GPS  & [10-12]  & 1.9 & 1.3 & 3.3 \\
\bottomrule
\end{tabular}
\end{center}
\vspace{-3mm}
\caption{Ablation study on generalized patch severance. }
\label{AbalationGPS}
\end{table}

\subsection{Generalization of Class-agnostic Mask Proposal Network} 
\label{supp:proposal}
To evaluate the generalization of the mask proposal network which is only trained on images belonging to seen classes, we report the recall of mask on COCO-Stuff dataset in Table~\ref{table:recallmask}. We calculated two metrics, $Recall@30$ and $Recall@50$. The results indicate that the mask proposal  network can provide class-agnostic masks for both seen and unseen classes. The network exhibits a satisfactory level of generalization performance.

\section{Visualization}
\paragraph{Heatmap visualization.} We show the visualization of heatmaps in Figure~\ref{figure:segment} which are supplementary cases for Section~\ref{sunsec:visualization} of the main paper. The self-attention in the transformer encoder layer retrieves information from a global scope, which may actually introduce more noise, and is therefore disadvantageous for the segmentation task.
$sheep$ class, $elephant$ class and $skateboard$ class are included in ``thing" category, and   $road$ are categorized as ``stuff" category. 
It demonstrates that our method can work well on both ``thing" and ``stuff" categories. CAL module is capable of guiding the model to focus on more distinctive regions, which can enhance its ability to accurately classify different categories.
\paragraph{Segmentation results.} 
As shown in Figure~\ref{fig:VisualSegment}, we present the visualized predictions of DeOP. Our method can effectively segment regions belonging to different categories. The method can achieve impressive results even when presented with regions of unseen categories (\textit{giraffe}, \textit{tree}, \textit{frisbee} and \textit{grass}), indicating its remarkable generalization performance and effectiveness for the zero-shot segmentation task.

\end{document}